\documentclass[runningheads]{llncs}

\usepackage{graphicx}
\usepackage{amsmath}
\usepackage{amssymb}
\usepackage{booktabs}
\usepackage{multirow}
\usepackage{cleveref}

\begin{document}
\title{Bounding and Filling: A Fast and Flexible Framework for Image Captioning}
%
%
\author{Zheng Ma\inst{1} \and
Changxin Wang \inst{1} \and
Bo Huang \inst{1} \and
Zixuan Zhu \inst{2} \and
Jianbing Zhang \inst{1} \thanks{Corresponding author.}
}
\authorrunning{Z. Ma et al.}
%
\institute{Nanjing University \and
University of Glasgow \\
\email{\{maz,cx.wang,191300018\}@smail.nju.edu.cn \\
zzx349313@gmail.com, zjb@nju.edu.cn
}
}

\maketitle              
\begin{abstract}
Most image captioning models following an autoregressive manner suffer from significant inference latency. 
Several models adopted a non-autoregressive manner to speed up the process. 
However, the vanilla non-autoregressive manner results in subpar performance, since it generates all words simultaneously, which fails to capture the relationships between words in a description. The semi-autoregressive manner employs a partially parallel method to preserve performance, but it sacrifices inference speed.
In this paper, we introduce a fast and flexible framework for image captioning called \textbf{BoFiCap} based on bounding and filling techniques. 
The BoFiCap model leverages the inherent characteristics of image captioning tasks to pre-define bounding boxes for image regions and their relationships. Subsequently, the BoFiCap model fills corresponding words in each box using two-generation manners. 
Leveraging the box hints, our filling process allows each word to better perceive other words.
Additionally, our model offers flexible image description generation: 1) by employing different generation manners based on speed or performance requirements, 2) producing varied sentences based on user-specified boxes.
Experimental evaluations on the MS-COCO benchmark dataset demonstrate that our framework in a non-autoregressive manner achieves the state-of-the-art on task-specific metric CIDEr (125.6) while speeding up 9.22× than the baseline model with an autoregressive manner; in a semi-autoregressive manner, our method reaches 128.4 on CIDEr while a 3.69× speedup.

\end{abstract}

\section{Introduction}
\label{sec:intro}

Image captioning tasks require models to automatically generate a sentence that describes a given image. 
Thanks to advancements in computer vision \cite{resnet,faster_rcnn} and natural language processing \cite{lstm,transformer}, numerous captioning models can accurately describe images \cite{bottom-up,cornia2020meshed,vinyals2015show}.
The application of image captioning extends to various fields, including facilitating language learning in children, aiding visually impaired individuals in comprehending their environment, and alerting drivers to potential hazards during autonomous driving.

\begin{figure}[t!]
    \setlength{\belowcaptionskip}{-0.5cm}
    \centering
    \includegraphics[width=0.6\textwidth]{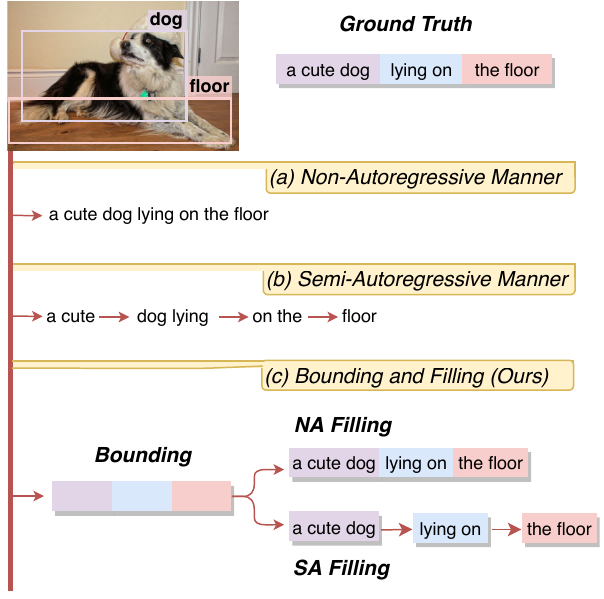}
    \caption{ (Top) exhibits a description that can be split into several boxes, each of which describes a region in the image or a relation between two regions.
    (Bottom) exhibits the difference between non-autoregressive manner, semi-autoregressive manner, and our BoFiCap. Our model bounds a series of boxes in advance, and then flexibly fills words in these boxes using NA or SA filling manner. }
    \label{fig:intro}
\end{figure}

In a real-time application, inference speed is a crucial factor to consider as models need to provide quick responses. 
However, many high-performing captioning models utilizing an autoregressive manner \cite{huang2019attention,LiPY022,SongZDT021} generate descriptions incrementally, word by word, resulting in significant inference latency. 
Researchers have recognized this challenge and have employed non-autoregressive \cite{fei2019fast,fei2020iterative,gao2019masked,guo2020non} or semi-autoregressive manner \cite{fei2021partially,zhou2021semi} to speed up.
The non-autoregressive manner generates a description in parallel, as displayed in \Cref{fig:intro}(a), which greatly improves the inference speed but often leads to performance degradation, repetition, and omissions \cite{fei2021partially}. These issues arise due to the absence of inter-word dependencies since each word is predicted simultaneously, leading to a limited understanding of its contextual role within the description.
The semi-autoregressive manner combines parallel and serial generation, as illustrated in \Cref{fig:intro}(b), striking a balance between speed and performance. However, it only produces a fixed and semantically meaningless chunk at each step. 

It is observed that descriptions typically employ a noun phrase to depict an image region and utilize a conjunctive or verb phrase to articulate the relationship between two regions.  As depicted in \Cref{fig:intro} (Top), \emph{`a cute dog'}, and \emph{`the floor'} are noun phrases, with each phrase describing a distinct region in the image. \emph{`lying on'} is a verb phrase that articulates the relationship between the two noun phrases.
If certain boxes representing regions or the relationship between regions can be predefined, the captioning model can generate descriptions based on the arrangement of these boxes. 
By doing so, words can establish mutual understanding through their association with specific boxes.
Drawing inspiration from this concept, in this paper we propose a novel framework for fast and flexible image captioning via bounding and filling named BoFiCap. 
To begin with, BoFiCap establishes a series of bounding boxes to facilitate description generation.
During the decoding phase, BoFiCap utilizes two filling methods, namely, non-autoregressive (NA) filling and semi-autoregressive (SA) filling, to populate the boxes with the appropriate words.
As illustrated in \Cref{fig:intro} (c), NA filling corresponds to the non-autoregressive manner, simultaneously populating all boxes, while SA filling aligns with the semi-autoregressive manner, progressively populating each box in sequential steps. 
Regarding the distinction in decoding between the two filling methods, SA filling exhibits the better ability to capture word dependencies compared to NA filling, as the SA method fills boxes incrementally, thus allowing it to leverage boxes already populated with words.
Hence, we introduce an imitation strategy wherein NA filling imitates the behavior of SA filling to improve its capability to comprehend other words.
Moreover, BoFiCap has the capability to generate diverse descriptions in a flexible manner by choosing various filling methods or providing different arrangements of boxes.
The main contributions of this paper are outlined as follows:
\begin{itemize}
    \item We propose BoFiCap, a fast and flexible framework for image captioning that decouples the generating process into bounding and filling stages.  BoFiCap utilizes advanced bounding techniques to define boxes and subsequently fills them using either the NA or SA filling approach. Additionally, our model offers the capability to generate diverse descriptions for the same image in a flexible manner.
    \item To enhance the performance of BoFiCap, we introduce parameter sharing between the decoders of the NA and SA methods. Furthermore, we propose an imitating strategy that improves the ability of NA filling to capture word dependencies.
    \item Experimental results demonstrate the effectiveness of our approach. In a non-autoregressive manner, our method achieves state-of-the-art performance while achieving a 9.22× speedup compared to the baseline model. In a semi-autoregressive manner, our method achieves a CIDEr score of 128.4, accompanied by a 3.69× speedup. 
\end{itemize}

\section{Preliminary}
\label{sec:prliminary}
In this section, we will briefly review three decoding manners in literature, including autoregressive manner, non-autoregressive manner, and semi-autoregressive manner. Given an image $I$, the image captioning task is requested to generate a description $S = \{ w_1, w_2, \dots, w_T \}$, with $T$ denoting the total length of this description.

\paragraph{\textbf{Autoregressive Manner.}}
In an autoregressive manner, the output of the $t$-th step depends on the previous sequence, and the probability of a sequence is the combination of the probabilities of all words. 
It can use the chain rule to decompose the likelihood of sequences:
\begin{align}
    P(S|I) = \prod_{t=1}^{T}P(w_{t}|w_{<t}, I), 
\end{align}
where $w_{<t} = \{ w_{1}, w_{2}, \dots, w_{t-1} \}$ represents the generated words before step $t$.

\paragraph{\textbf{Non-Autoregressive Manner.}}
The non-autoregressive manner is proposed to address the high inference latency problem.
It breaks dependency on previously generated words and generates them in parallel, which can be formulated as:
\begin{align}
    P(S|I) = \prod_{t=1}^{T}P(w_{t}|I). 
\end{align}

\paragraph{\textbf{Semi-Autoregressive Manner.}}
\label{sec:back_semi}

In a semi-autoregressive manner, a group of words is generated in one step, which is parallel in the group and autoregressive between groups. Assuming $S$ can be divided in $\{g_1, g_2, \dots, g_N \}$, where $N$ is the number of groups, the semi-autoregressive manner can be formulated as:
\begin{align}
    P(S|I) = \prod_{t=1}^{N}P(g_{t}|g_{<t}, I),
    \label{eq:semi}
\end{align}
where $g_{<t} =\{ g_1, g_2, \dots, g_{t-1} \}$ is the generated groups before step t.

\section{Proposed Method}

\vspace{-0.8cm}
\begin{figure}[h!]
    \centering
    \includegraphics[width=0.7\textwidth]{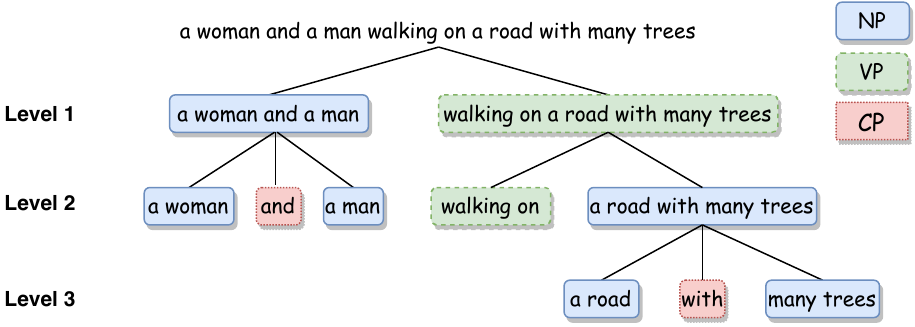}
    \caption{A description is parsed into a tree structure by a constituency parser. 
    }
    \label{fig:split}
\end{figure}
\vspace{-0.8cm}

\subsection{Generating Hierarchical Boxes}
\label{sec:split}

We use a constituency parser \cite{DBLP:journals/corr/abs-2007-14640} to split sentences in advance. 
To be specific, a description is parsed into a hierarchical tree structure in which the layers from shallow to deep represent coarse-to-fine bounding information. 
As shown in \Cref{fig:split}, a whole description is first divided into  
\emph{`a woman and a man'} and \emph{`walking on a road with many trees'}. Then, the two items can be divided into finer components.
We define three types of boxes: NP-box, VP-box, and CP-box, corresponding to NP, VP, and CP labels in the constituency parser
We name them level-$k$ to distinguish different levels, where $k=-1$ represents all phrases that can no longer be cut.

\vspace{-0.6cm}
\begin{figure*}[h!]
    \setlength{\belowcaptionskip}{-1.0cm}
    \centering
    \includegraphics[width=\textwidth]{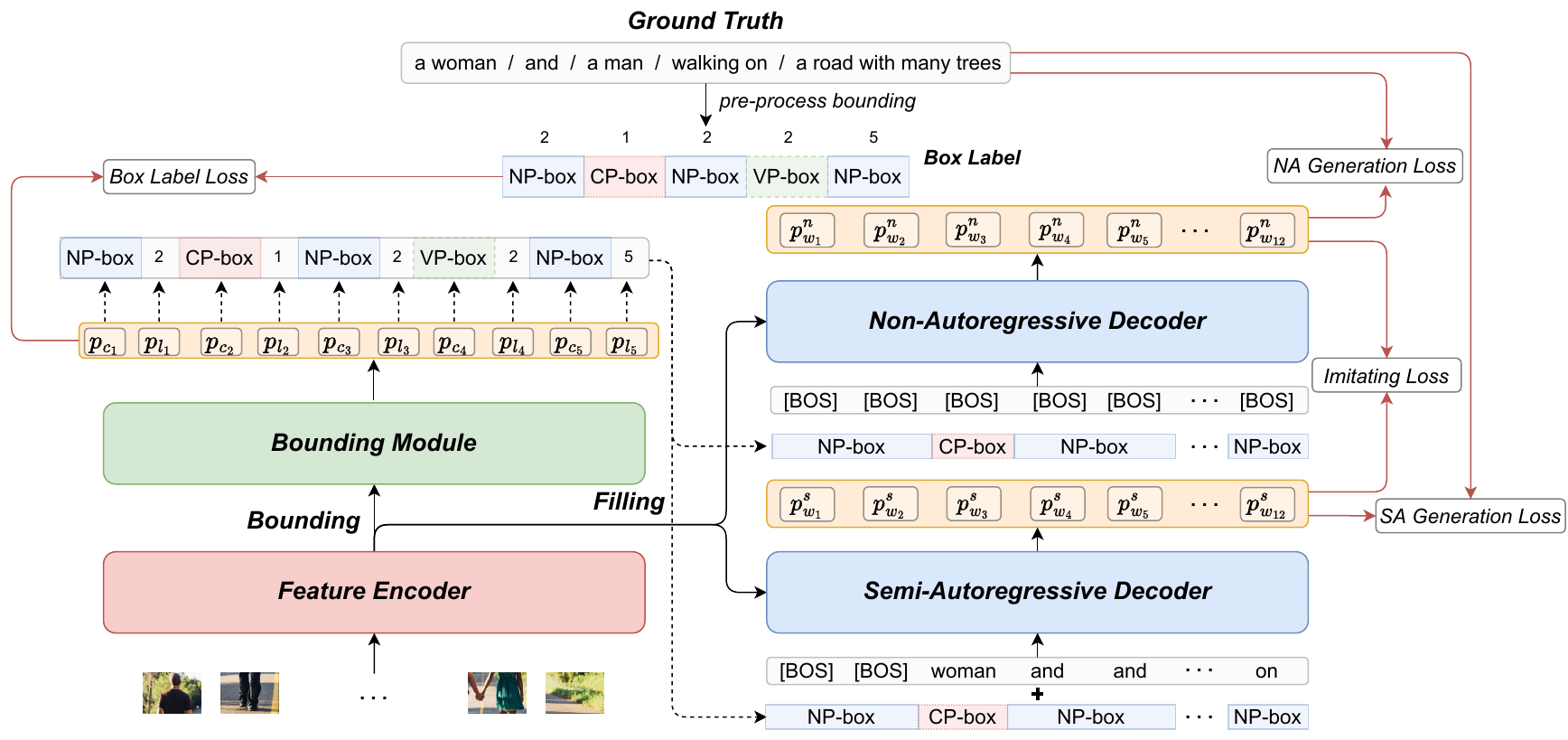}
    \caption{The illustration of our BoFiCap. (Low Left) exhibits the regions are encoded by the feature encoder. (Upper Left) exhibits the process of bounding, it will predict bounding boxes information for filling. (Right) exhibits our BoFiCap for NA and SA filling manners. Noted that the non-autoregressive and semi-autoregressive decoder share parameters. }
    \label{fig:arch}
\end{figure*}

\subsection{BoFiCap Model Architecture}
\label{sec:model}
Generally, our BoFiCap model is built on Transformer \cite{transformer}. As displayed in \Cref{fig:arch}, it consists of three modules: feature encoder, bounding module, and filling module.

\paragraph{\textbf{Feature Encoder.}}
The feature encoder aims to encode the image into a visual context. In our framework, we extract regions (denoted as $R$) from raw images in advance by a faster-RCNN model \cite{bottom-up} which uses bottom-up attention and is trained on Visual Genome datasets \cite{vg}. 
The architecture of the feature encoder is as same as the encoder of the vanilla Transformer \cite{transformer}, but we remove the position encoding because regions cannot be arranged linearly.

\paragraph{\textbf{Bounding Module.}}
To bound a series of boxes, three aspects should be considered in our bounding module: 1) the number of boxes, denoted as $N$. 2) the number of words in each box, denoted as $ L=\{ l_{1}, l_{2}, \dots, l_{N} \} $. 3) the type of each box, denoted as $ C=\{ c_{1}, c_{2}, \dots, c_{N} \} $. Further, we can define bounding information as $ B = \{ b_{1}, b_{2}, \dots, b_{N} \} $, where $b_{i} = \{ \underbrace{c_{i}, c_{i}, \dots, c_{i}}_{l_{i}} \}$, thus enabling $B$ to integrate the above three aspects of information.

Our bounding module generates bounding information autoregressively, which consists of one Transformer decoder layer and two linear classifiers to predict the type of boxes and the number of corresponding words.
Specifically, conditioning on the image regions extracted from raw images, our bounding module predicts the bounding information of one box, namely the type of one box and the number of corresponding words, in one step. The probability of bounding information can be factorized as:
\begin{align}
    P(B|R) = \prod_{t=1}^{N}P(b_{t}|b_{<t}, R) = \prod_{t=1}^{N}P(l_{t}, c_{t}|b_{<t}, R). 
\end{align}

\paragraph{\textbf{Filling Module.}}
Our filling module includes two decoders with different manners: the non-autoregressive decoder and semi-autoregressive decoder for NA and SA filling respectively, which are as same as the decoder of the vanilla Transformer architecturally.
Three pieces of information are input in our filling module, including visual context, bounding information and history words. 

In SA filling manner, the SA decoder will fill one box with words in one step. 
The target description probability can be factorized as:
\begin{align}
    P(S|B,R) = \prod_{t=1}^{N}P(g_{t}|g_{<t}, b_{\leq t}, R),
\end{align}
where $g_{t}$ represents the filled words in box $b_{t}$. Note that the number of words in each box $b_{i}$ may be different, which means that the generated words at the $t-1$-th step can not be directly regarded as the input at the $t$-th step.
To address this issue, we propose a copy strategy called Position-wise Copy. Intuitively, the two words that are closer in position are likely to be closer in relation. Assuming the output of last step is $g_{t-1} = \{w_1, w_2, \dots, w_{l_{t-1}} \}$, and current step's output is $g_{t} = \{w_1^{'}, w_2^{'}, \dots, w_{l_{t}}^{'} \}$, each word $w_{i}$ in $g_{t-1}$ will be copied $n_{i}$ times by Position-wise Copy:
\begin{align}
n_{i} = 
\left\{ \begin{matrix}
\lfloor l_{t}/l_{t-1} \rfloor, & i \leq l_{t-1} - l_{t} \% l_{t-1}, \\ 
\lfloor l_{t}/l_{t-1} \rfloor + 1, & i > l_{t-1} - l_{t} \% l_{t-1},
\end{matrix} \right.  
\end{align}
where $\lfloor \cdot \rfloor$ is a floor function.

In the NA filling manner, the NA decoder fills all boxes at once. The target description probability can be factorized as:
\begin{align}
    P(S|B,R) = \prod_{t=1}^{T}P(w_{t}|B, R).  
\end{align}

\vspace{-.8cm}
\subsection{Imitating Strategy}
\label{sec:imitating}
Because the SA filling is partially parallel, which fills a box in one step, it better captures the relationship between words and performs better than the NA filling. 
Therefore, in our model we let the NA filling imitate the SA filling by an online knowledge distillation method \cite{DBLP:conf/cvpr/ZhangXHL18}. We follow the previous work \cite{DBLP:journals/corr/HintonVD15} to make the output of the NA filling close to the SA filling when training them jointly.

In the filling stage, assuming the target words' probabilities of NA filling and SA filling are $\{ p^n_{w_1}, p^n_{w_2}, \dots, p^n_{w_T} \}$ and $\{ p^s_{w_1}, p^s_{w_2}, \dots, p^s_{w_T} \}$ respectively, the imitating loss can be represented by a Kullback-Leibler (KL) Divergence as:
\begin{align}
    &\mathcal{L}_{Imit} = \frac{1}{T} \sum_{t=1}^{T} p^n_{w_t} \log{\frac{p^n_{w_t}}{p^s_{w_t}}}. 
\end{align}

\vspace{-.8cm}
\subsection{Model Training}
\label{sec:train}

\paragraph{\textbf{CE Training Stage.}}
Our BoFiCap model contains two steps in the CE training stage: bounding and filling. The bounding step needs to predict the type of boxes and the number of words to be filled in each box.
The objective of this step is to minimize the negative log-likelihood of the correct
type of boxes and the number of words using the maximum likelihood estimation (Box Label Loss):
\begin{align}
    \mathcal{L}_{Bound} = -\sum_{t=1}^{N} \log P_{\theta}(l_{t}, c_{t}|b_{<t}, R). 
\end{align}

The filling step is filling all boxes predicted in advance using NA or SA filling.
NA filling fills the boxes in parallel, the loss of which is the sum of the negative log-likelihood of the correct words (NA Generation Loss) as:
\begin{align}
\mathcal{L}_{NA} = -\sum_{t=1}^{T} \log P_{\theta}(w_{t}|B, R).
\end{align}

SA filling fills a box in one step, the loss of which can be calculated (SA Generation Loss) as:
\begin{align}
    \mathcal{L}_{SA} = -\sum_{t=1}^{N} \log P_{\theta}(g_{t}|g_{<t}, b_{\leq t}, R). 
\end{align}

We train our BoFicap model jointly in NA and SA filling, so the objective function for BoFiCap can be written as:
\begin{align}
    \mathcal{L} = \mathcal{L}_{Bound} + \mathcal{L}_{NA} + \mathcal{L}_{SA} + \mathcal{L}_{Imit}.
\end{align}

\paragraph{\textbf{RL Training Stage.}}
In the RL training stage, we directly minimize the negative expected reward using SCST \cite{rennie2017self}:
\begin{align}
    \mathcal{L}= - \frac{1}{M} \sum_{m=1}^{M}(r(S_n)-b) \bigtriangledown \log(p_{\theta}(S_n)), 
\end{align}
where $r$ is the CIDEr metric, $p_{\theta}(S_n)$ is the probability of the $m$-th sample description, and $b$ is the baseline reward. We refer to the previous work \cite{DBLP:journals/corr/abs-2003-09971} for the value of b and set $M=5$ in our experiments.

\begin{table*}
    \setlength{\belowcaptionskip}{-0.2cm}
    \centering
    \footnotesize
    \resizebox{0.98\textwidth}{!}{
    \begin{tabular}{l  c c c c c c  c c}
    \toprule
        Models & BLEU-1 & BLEU-4 & METEOR & ROUGE & CIDEr & SPICE & Latency & Speedup \\
    \midrule 
    \midrule
        \multicolumn{9}{l}{\textit{Autoregressive Models}}\\
    \midrule
        AIC (beam=1) & 80.5 & 38.9 & 29.0 & 58.7 & 129.4 & \textbf{22.8} & 192ms & \textbf{1.73×}\\
        AIC (beam=3) & \textbf{80.9} & \textbf{39.3} & 29.0 & \textbf{58.9} & 130.2 & \textbf{22.8} & 332ms & 1.00×\\
    \midrule
        \multicolumn{9}{l}{\textit{Non-Autoregressive Models}}\\
    \midrule
        MNIC \cite{gao2019masked} & 75.4 & 30.9 & 27.5 & 55.6 & 108.1 & 21.0 & - & 2.80× \\
        FNIC \cite{fei2019fast} & / & 36.2 & 27.1 & 55.3 & 115.7 & 20.2 & - & 8.15× \\
        CMAL \cite{guo2020non} & \textbf{80.3} & 37.3 & 28.1 & 58.0 & 124.0 & 21.8 & - & \textbf{13.90×}\\
        IBM  \cite{fei2020iterative} & 77.2 & 36.6 & 27.8 & 56.2 & 113.2& 20.9 & - & 3.06×\\\midrule
        BoFiCap-NA (\textit{Ours}) & 80.1 & \textbf{38.2} & \textbf{28.4} & \textbf{58.2} & \textbf{125.6} & \textbf{22.1} & 36ms & 9.22×\\
    \midrule
        \multicolumn{9}{l}{\textit{Semi-Autoregressive Models}}\\
    \midrule
        PNAIC \cite{fei2021partially} & 80.4 & 38.3 & 29.0 & 58.4 & \textbf{129.4} & 22.2 & - & 2.17×\\
        SATIC \cite{zhou2021semi} & \textbf{80.8} & 38.4 & 28.8 & 58.5 & 129.0 & \textbf{22.7} & - & 1.65×\\
        SAIC \cite{yan2021semi} & 80.4 & 38.7 & \textbf{29.4} & 58.5 & 128.3 & 22.2 & - & 1.55×\\\midrule
        BoFiCap-SA (\textit{Ours}) & 80.5 & \textbf{38.9} & 28.8 & \textbf{58.8} & 128.4 & \textbf{22.7} & 90ms & \textbf{3.69×}\\
    \bottomrule
    \end{tabular}
    }
    \caption{Performance comparisons with different evaluation metrics on the test set of MS COCO. All values except Latency and Speedup are reported as a percentage (\%). `/' denotes that the results are not reported.`-' denotes unfair comparison because latency is greatly affected by different devices so we do not report them in other models. Our Latency is tested on a GeForce GTX 1080 Ti GPU. The Speedup values are from the corresponding papers. The top results under each decoding manner are in bold. BoFiCap-NA and BoFiCap-SA are our models with the non-autoregressive filling and the semi-autoregressive filling, respectively.}
    \label{tab:main}
\end{table*}
\section{Experiments}

\subsection{Experimental Settings}
We experiment with the MS COCO dataset \cite{chen2015microsoft} that is widely used in image captioning tasks, including 123,287 images and each image has five captions at least.
We refer to the karpathy's split \cite{karpathy2015deep} to split the dataset into 113,287, 5,000, and 5,000 images for training, validation, and offline testing.
We count all words in captions and omit words that occur less than five times to build a vocabulary that comprises 9,487 words. In addition, for efficiency, we truncate captions that are longer than 16 in training. We use the autoregressive captioning model (AIC) as our teacher model.

\paragraph{\textbf{Evaluation Metrics.}}

We evaluate our model using a variety of automatic metrics, including BLEU-1/4 \cite{papineni2002bleu}, METEOR \cite{banerjee2005meteor}, ROUGE \cite{lin2004rouge}, CIDEr \cite{vedantam2015cider} and SPICE \cite{anderson2016spice}, which denoted as B1/4, M, R, C, and S, respectively for convenience.
We also use latency and speedup to compare the efficiency to other models. 

\subsection{Main Comparison Results}
We compare the performance of our proposed method to various baseline models in two manners and report the results in \Cref{tab:main}. Our observations are summarized below.

Our proposed method BoFiCap-NA outperforms a variety of compelling non-autoregressive baselines in NA filling, regarding BLEU-4, METEOR, ROUGE, CIDEr, and SPICE metrics, while maintaining a 9.22× speedup.
In SA filling, we observe that our proposed method BoFiCap-SA has a competitive performance compared to models in a semi-autoregressive manner, which achieves state-of-the-art results on BLEU-4, ROUGE, and SPICE while maintaining a 3.69× speedup. 
We report the result of the BoFiCap-SA trained solely since joint training has a negative impact on its performance, and we will discuss this phenomenon in \Cref{sec:ablation_study}.

\vspace{-.5cm}
\section{Analysis}
\vspace{-.3cm}

\paragraph{\textbf{Effect of Hierarchical Split Method.}}
\label{sec:effect_level}
Performance and speed evaluation of BoFiCap models at different hierarchical splits level-$k$ is reported in \Cref{tab:hier_dataset}. 
We observe that in SA filling, the acceleration primarily results from the reduction of decoding steps because the model trained with a shallow layer split generates descriptions with fewer boxes. Comparing the results from layers $1$ to $2$, we see that the speedup changes from 5.19× to 8.51× while only suffering a slight drop in the CIDEr score (1.6 points).
In NA filling, the acceleration mainly results from the reduction of bounding box steps and predicting the number of corresponding words because our bounding module bounds boxes in an autoregressive manner. Comparing the results from layers $1$ to $2$, we see that the speedup changes from 11.45× to 13.83× while only experiencing a slight drop in the CIDEr score (1.7 points).
\vspace{-.7cm}
\begin{table}
    \centering
    \small
    \tabcolsep=4pt
    \begin{tabular}{l c|c c c c c c}
    \toprule
        Models & $k$ & B4 & M & R & C & S & Speedup\\
    \midrule
    \midrule
        AIC(beam=3) & / & 39.3 & 29.0 & 58.9 & 130.2 & 22.8 & 1.00× \\
    \midrule
        \multirow{3}{*}{BoFiCap-NA} & 1 & 36.8 & 27.9 & 57.7 & 122.1 & 21.5 & 13.83×  \\
                                 & 2 & 37.6 & 28.2 & 58.1 & 123.8 & 21.9 & 11.45×  \\
                                 & -1 & 38.2 & 28.4 & 58.2 & 125.6 & 22.1 & 9.22× \\ 
    \midrule
        \multirow{3}{*}{BoFiCap-SA} & 1 & 37.6 & 28.1 & 58.2 & 123.3 & 21.7 & 8.51× \\
                                 & 2 & 38.2 & 28.4 & 58.4 & 124.9 & 22.2 & 5.19× \\
                                 & -1 & 38.5 & 28.7 & 58.5 & 127.5 & 22.6 & 3.69× \\
    \bottomrule
    \end{tabular}
    \caption{Effect of hierarchical box splits evaluated on MSCOCO test set. $k$ denotes the level of box splits.}
    \label{tab:hier_dataset}
\end{table}
\vspace{-1.3cm}

\paragraph{\textbf{Ablation Studies.}}
\label{sec:ablation_study}

In a non-autoregressive manner, we implement a single vanilla non-autoregressive model named NAIC. 
In \Cref{tab:ablation}, we observe that our BoFiCap-NA with bounding boxes greatly improves the CIDEr score from 108.5 to 122.3. 
Further improvements can be achieved by adding joint training and imitating strategy, resulting in our BoFiCap-NA model reaching the state of the art on the CIDEr score (125.6).
Regarding our semi-autoregressive manner, the joint training method results in a slight drop in the performance of our BoFiCap-SA. This is because we use a shared decoder to help our BoFiCap-NA, and their decoding processes are different. We also observe that the imitating strategy hardly works in BoFiCap-SA because it allows the BoFiCap-NA model to imitate the output of the BoFiCap-SA model, but this strategy has little effect on BoFiCap-SA.
\begin{table}
    \centering
    \small
    \tabcolsep=4.5pt
    \begin{tabular}{l l|c c c c c}
    \toprule
        Models & Methods & B4 & M & R & C & S\\
    \midrule
    \midrule
        NAIC & / & 30.8 & 25.6 & 55.4 & 108.5 & 19.6\\
    \midrule
        \multirow{3}{*}{BoFiCap-NA} & +Bound & 37.6 & 28.1 & 58.0 & 122.3 & 21.8 \\
                                 & +Bound+Joint & 37.9 & 28.3 & 58.2 & 124.7 & 22.0 \\
                                 & +Bound+Joint+Imit & 38.2 & 28.4 & 58.2 & 125.6 & 22.1 \\
    \midrule
        \multirow{3}{*}{BoFiCap-SA} & +Bound & 38.9 & 28.8 & 58.8 & 128.4 & 22.7 \\
                                 & +Bound+Joint & 38.6 & 28.7 & 58.5 & 127.2 & 22.7 \\
                                 & +Bound+Joint+Imit & 38.5 & 28.7 & 58.5 & 127.5 & 22.6 \\
    \bottomrule
    \end{tabular}
    \caption{Effect of three methods evaluated on MS COCO test set, where Bound, Joint, and Imit represent the bounding boxes, joint training, and Imitating strategy.}
    \label{tab:ablation}
\end{table}
\vspace{-1.2cm}

\begin{figure*}[h!]
    \centering
    \includegraphics[width=\textwidth]{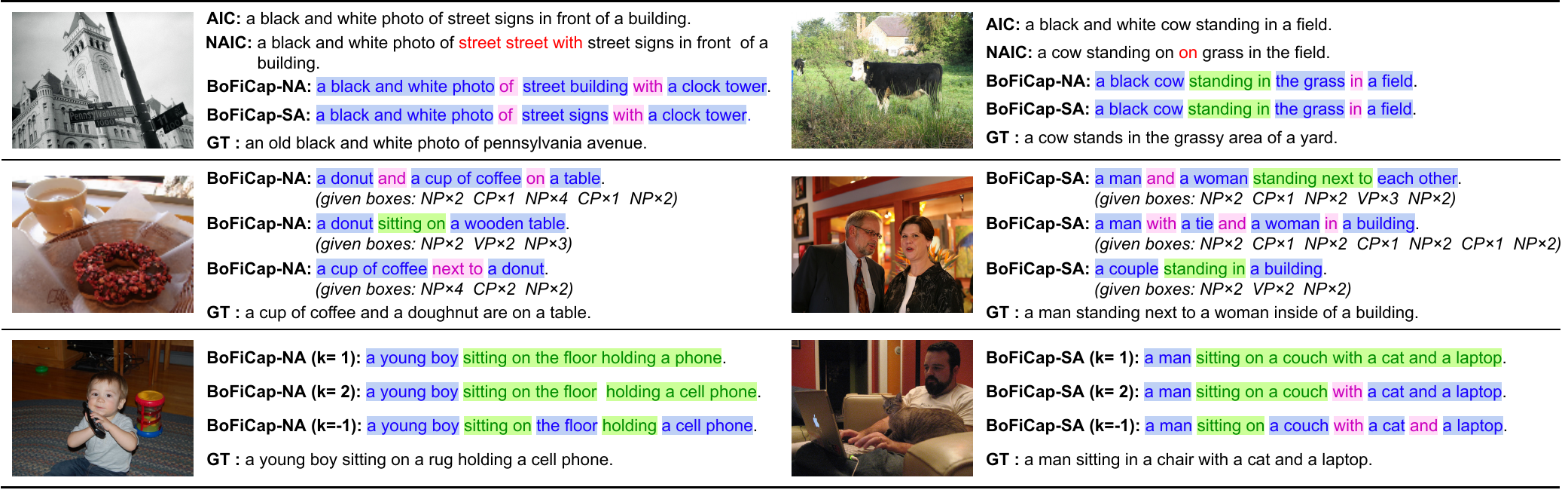}
    \caption{Examples of captions generated from different models. GT denotes ground-truth captions. }
    \label{fig:case_study}
\end{figure*}

\paragraph{\textbf{Case Study.}}
\label{sec:case_study}
We present six examples in \Cref{fig:case_study}.
In the top two examples shown in \Cref{fig:case_study}, we compare the sentences generated from AIC, NAIC, and two manners of BoFiCap models.
Overall, all the models effectively represent the visual content of the given image.
Compared to the baseline NAIC, the sentences generated by both manners of our BoFiCap model are fluent and precise. Moreover,our BoFiCap model can accurately assign words with the corresponding box type, resulting in descriptions that are more syntactically structured and less repetitive. To exhibit the diverse generation ability of BoFiCap, we also provide multiple sentences generated with different boxes or split levels in the middle and bottom of \Cref{fig:case_study}.

\vspace{-0.5cm}
\section{Related Work}
\vspace{-0.3cm}
\paragraph{\textbf{Non-Autoregressive Image Captioning.}}
Overall, non-autoregressive image captioning models can be divided into two categories: latent transformer \cite{kaiser2018fast} based and iterative refinement based. 
Following the latent transformer, Fei\cite{fei2019fast} firstly predicts ordered keywords with an RNN and then generates the complete sentence simultaneously. Gao \textit{et al.}\cite{gao2019masked} utilize a refinement strategy on sentence generation, which means the sentence is iterated multiple times and each iteration generates part of the words in the final sentence. Fei \cite{fei2020iterative} also adopts an iterative refinement strategy and constructs a latent variable to bridge the image encoder and textual decoder, which is iteratively optimized to improve generation quality during inference. Besides, Guo \textit{et al.}\cite{guo2020non} propose to use multi-agent reinforcement learning to model the sentence-level objective. Different from all the above work, our NA filling method makes two innovations: 1) we use the bounding boxes to enhance the decoder's ability to capture dependencies between words. 2) we improve the performance of the NA filling method by jointly training the model with two filling manners and using the imitating strategy.

\paragraph{\textbf{Semi-Autoregressive Image Captioning.}}

Although the performance of the non-autoregressive model has been enhanced by many methods, it still falls behind of state of the art results \cite{cornia2020meshed,huang2019attention}. Recently, some works have explored how to make a trade-off between quality and speed by utilizing a semi-autoregressive manner.
Generally, these methods generate descriptions in a group type introduced in \Cref{sec:back_semi}.
Zhou \textit{et al.}\cite{zhou2021semi} simply treat each block as one group. 
Fei\cite{fei2021partially} organizes the words in the same position in each block into one group. 
Yan \textit{et al.}\cite{yan2021semi} adopt a slightly special two-stage generation paradigm, where the first word in each block is generated autoregressively, then the rest words will be generated simultaneously. 
But our groups are meaningful phrases rather than fixed blocks, and we utilize bounding boxes as hints for each group generation.

\vspace{-0.5cm}
\section{Conclusion}
\vspace{-0.3cm}
In this paper, we propose a new framework for image captioning, BoFiCap, that utilizes bounding and filling techniques. In contrast to previous accelerated approaches, our method utilizes the properties of descriptive sentences to decompose generation steps using bounding and filling. Furthermore, our framework provides flexible image description generation to meet the specific needs of users. 

\vspace{-0.5cm}
\section{Acknowledgements}
\vspace{-0.3cm}
We would like to thank the anonymous reviewers for their constructive comments. This work was supported by NSFC No. 62176115.
\vspace{-0.5cm}

%
%
\bibliographystyle{splncs04}
\bibliography{llncs2e/ref}
%

\end{document}